\documentclass{article}

\PassOptionsToPackage{numbers,compress}{natbib}
\usepackage[dblblindworkshop,final]{neurips_2026}
\usepackage{amsmath}
\usepackage{float}
\usepackage{rotating}
\workshoptitle{Foundation Models for Temporal Systems: From Forecasting to World Modeling}
\usepackage[utf8]{inputenc} 
\usepackage[T1]{fontenc}    
\usepackage{hyperref}       
\usepackage{url}            
\usepackage{booktabs}       
\usepackage{amsfonts}       
\usepackage{nicefrac}       
\usepackage{microtype}      
\usepackage{graphicx}
\usepackage{xcolor}         
\hypersetup{
    colorlinks=false,
    linkbordercolor={1 0 0},
    citebordercolor={0 1 0},
    urlbordercolor={0 0 1},
    pdfborder={0 1 1}
}
\title{ST-GNN-AAAS: Spatio-Temporal Graph Neural Networks for Anytime Automatic Algorithm Selection in the Traveling Salesman Problem}

\author{%
  Zeyu Zhang \\
  Division of Natural and Applied Sciences \\
  Duke Kunshan University \\
  Kunshan, China \\
  \texttt{zeyu.zhang@dukekunshan.edu.cn} \\
  \And
  Mustafa Misir \\
  Division of Natural and Applied Sciences \\
  Duke Kunshan University \\
  Kunshan, China \\
  \texttt{mustafa.misir@dukekunshan.edu.cn} \\
}

\begin{document}

\maketitle

\begin{abstract}
  Anytime Automatic Algorithm Selection (AAAS) aims to make solver choices under varying time budgets.
  The existing AAAS approaches largely rely on static instance characteristics and make limited use of temporal information revealed during solver execution. 
  We propose ST-GNN-AAAS, a Spatio-Temporal Graph Neural Network framework for AAAS in the Traveling Salesman Problem (TSP). The framework represents evolving incumbent tours using time-varying node features, including distances from each node to its neighbors in the current tour, and combines these features with solver performance observations over a short history window. Given this short history, ST-GNN-AAAS predicts solver selections over a longer future horizon. We evaluate the method on a subset of diverse TSP benchmarks from NATIONAL, TNM, TSPLIB, and VLSI, releasing the corresponding solver execution data to support reproducibility. Our model achieves a selection accuracy of 76.59\% and closes 46.05\% of the average rank gap between the single best solver (SBS) and the virtual best solver (VBS). 
\end{abstract}

\section{Introduction}

The Traveling Salesman Problem (TSP) is a well-studied NP-hard problem in combinatorial optimization, with applications in routing, logistics, scheduling and related planning tasks \cite{applegate2011traveling,baldacci2012new}. 
Over the years, numerous high-performance heuristic algorithms have been developed such as 
Chained Lin-Kernighan (CLK) \cite{applegate2003chained}, 
Lin-Kernighan-Helsgaun (LKH) \cite{helsgaun2000effective}, 
Edge Assembly Crossover (EAX) \cite{nagata2013powerful} and 
the Multiagent Optimization System (MAOS) \cite{xie2008multiagent}. 
However, the performance of these solvers varies substantially across different problem instances as in many other application domains \citep{kerschke2019automated}. 
Automatic Algorithm Selection (AAS) addresses this challenge by identifying the most suitable solver for a given instance rather than relying on a single algorithm for all cases \citep{rice1976algorithm}.
Standard AAS studies typically assume a fixed runtime budget. 
 

Existing AAAS methods, however, mainly rely on static instance representations or domain-specific handcrafted features. 
Static features have been used for knapsack \citep{Huerta2020knapsack}, 
structural instance features for multi-agent path finding \citep{Zapata_Godoy_Asin-Acha_2024}, 
domain-specific features for pseudo-Boolean optimization \citep{pezo2025automatic}, and coordinates of cities for the TSP \cite{huerta2022improving}. 
While these features characterize the problem instances, they capture limited information generated during solver execution.  
For the TSP in particular, the state of a solver evolves over time not only in its incumbent objective value but also in the structure of the incumbent tour. 
Each new incumbent solution may change the tour permutation and the neighboring cities associated with each node. As illustrated in Figure~\ref{fig:solver_dynamics} in Appendix~\ref{solver_dynamics}, the leading solver can also vary considerably across time steps. This motivates explicitly modeling the temporal information contained in solver trajectories together with the spatial structure of the TSP instance.


\noindent\textbf{Contributions:}
(1) We introduce ST-GNN-AAAS, a spatio-temporal framework for anytime automatic algorithm selection in TSP that combines the spatial structure of problem instances with temporal information from solver trajectories, enabling solver prediction over a long future horizon from a short observation window;
(2) we augment conventional solver execution records with incumbent tour permutations and derive time-varying node-level features from the evolving tour structure, while extending the solver portfolio to seven TSP algorithms;
(3) we evaluate the proposed framework using average rank as the primary metric, together with selection accuracy, SBS--VBS gap closure, and statistical significance tests, providing a more complete view of selection performance across runtime budgets.

\section{ST-GNN-AAAS}

\subsection{Problem Formulation}
\label{Sec2.1 Problem Formulation}
To address the anytime automatic algorithm selection using a Spatio-temporal Graph Neural Network, following Huerta et al.\cite{huerta2022improving}'s temporal discretization of time steps, solver trajectories were resampled onto a common exponential time grid of 500 time steps,which comprise $\mathcal{T}_{\mathrm{obs}} \text{ and } \mathcal{T}_{\mathrm{pred}}$. As the exact discretization formula was not provided in the original study, we reconstructed the time grid to match the reported range and temporal spacing, which spans \(t_{\min}=0.004\,\text{s}\) to \(t_{\max}=7200\,\text{s}\), given by
\begin{equation}
    t_s = t_{\min} + (t_{\max}-t_{\min}) \frac{e^{s/\tau}-1}{e^{(S-1)/\tau}-1}
\label{eq:time_grid}
\end{equation}
with \(\tau = 48.11\) and \(S=500\).
We then define the learning task of ST-GNN-AAAS over a historical observation window and a subsequent prediction horizon. We consider an observation window
$\mathcal{T}_{\mathrm{obs}}=\{t_s,\ldots,t_{s+L-1}\}$,
followed by a prediction horizon
$\mathcal{T}_{\mathrm{pred}}=\{t_{s+L},\ldots,t_{s+L+P-1}\}$,
where $s$ denotes the starting index step, $L$ the observation length,
and $P$ the prediction horizon. For each instance $i \in \mathcal{I}$, solver $a \in \mathcal{A}$,
and time point $t \in \mathcal{T}_{\mathrm{obs}}$, we denote the aggregated observations over the whole observation window for all solvers in the portfolio by
\[
\mathcal{X}_i^{\mathrm{obs}}
=
\{\mathbf{X}_{i,a,t}\mid a\in\mathcal{A},\, t\in\mathcal{T}_{\mathrm{obs}}\}.
\]
Given $\mathcal{X}_{i}^{\mathrm{obs}}$, the task is to predict a solver
for each time point in $\mathcal{T}_{\mathrm{pred}}$:
\begin{equation}
F_{\theta}(\mathcal{X}_{i}^{\mathrm{obs}})
\rightarrow
\{\hat{S}(i,t)\in\mathcal{A}
\mid t\in\mathcal{T}_{\mathrm{pred}}\}.
\label{eq:stgnn_task}
\end{equation}

\subsection{Spatial and Temporal Features}
\label{Sec2.2}

We represent each TSP instance as a graph, where cities are nodes and graph connectivity is constructed using a $k$-nearest-neighbor (KNN) graph with $k=10$. The city coordinates provide the static spatial information used by the graph encoder.

The incumbent tour changes during solver execution, so we additionally construct time-varying node features from the current tour structure. For each city, we record its distance to the predecessor and successor in the incumbent tour at each time step, denoted as $d_{\mathrm{prev}}$ and $d_{\mathrm{next}}$. These features change whenever the tour permutation changes and therefore provide a node-level description of how the incumbent solution evolves over time. 

We also use solver-level temporal features derived from the recorded objective trajectories. At each observed time step, these include normalized objective value, whether the solver matches the current virtual best solver (\texttt{is\_vbs}), normalized cost gap to current best solver, normalized solver rank (\texttt{rank\_norm}), and whether the incumbent objective has improved (\texttt{is\_improving}). All of these quantities are computed only from solver observations available within the history window.

\subsection{Model Architecture}

\begin{figure}[t]
    \centering
    \includegraphics[width=0.95\linewidth]{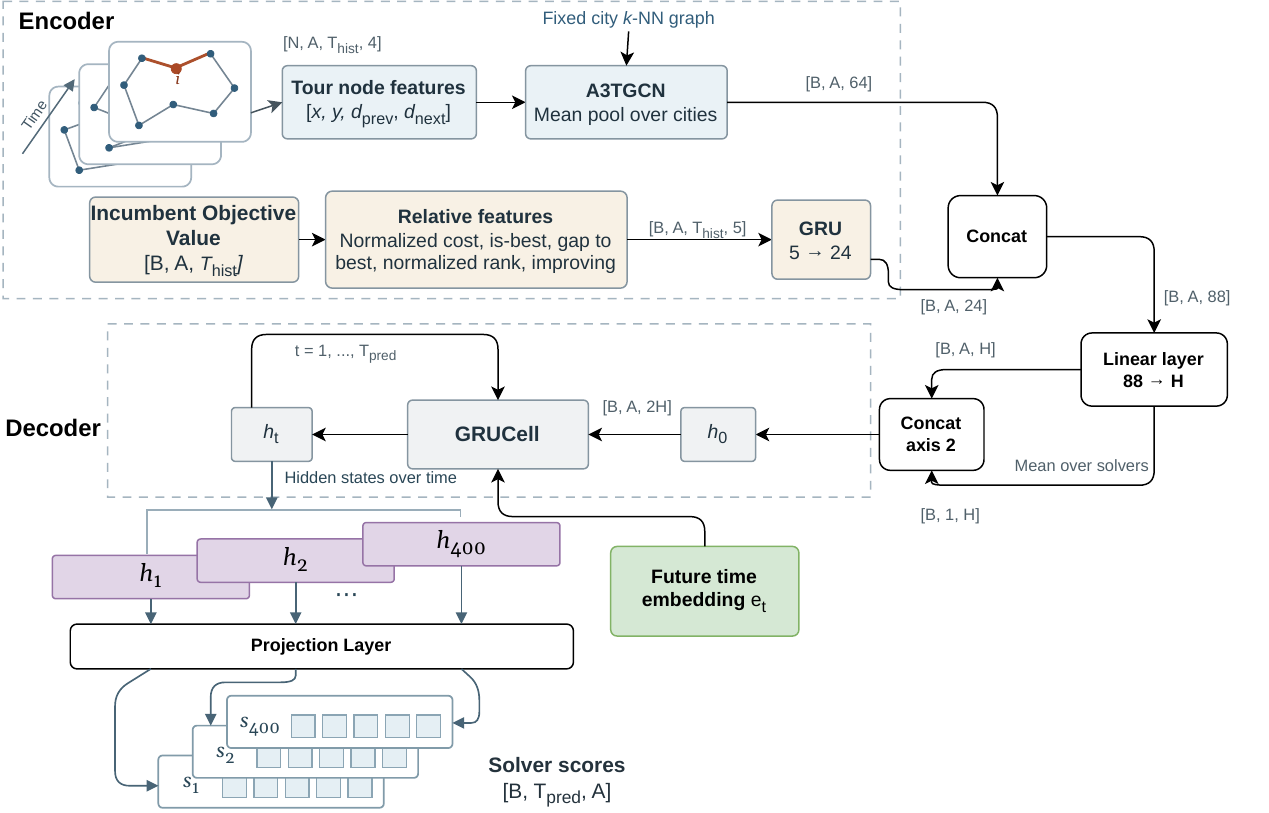}
    \caption{Overview of the ST-GNN-AAAS architecture. The model encodes
    spatial TSP structure together with temporal solver observations over
    the history window and predicts solver selections over the future
    horizon.}
    \label{fig:arch}
\end{figure}

Inspired by the encoder--decoder architecture of DCRNN
\cite{li2017diffusion}, our model encodes the observed spatio-temporal
solver states and predicts solver selections over a longer future
horizon. For the spatio-temporal encoder, we use A3T-GCN \cite{bai2021a3t}, paired with a GRU-based module. To train instances with varying sizes, code is implemented using PyTorch Geometric\cite{fey2019fast}. An overview of ST-GNN-AAAS is shown in
Figure~\ref{fig:arch}.

\section{Methods}

\subsection{Experimental Setup}

\textbf{Benchmark and solver runs.}
We use the TSP benchmark instances released by Huerta et al.~\cite{huerta2022improving} and retain instances with fewer than 5,000 cities. Details of the training and test sets are provided in \ref{dataset}. Each instance is executed with all seven solvers under a 2-hour wall-clock limit, with early termination if a better solution does not improve for 10 minutes. Solver-internal iteration limits are set sufficiently high so that termination is effectively determined by these time-based conditions. We make only minimal changes to the original implementations and log every incumbent improvement together with its execution time, objective value, and tour permutation.

\textbf{Main prediction setting.}
 We expand Huerta et al.'s five-solver portfolio to consider seven TSP solvers including  Chained Lin-Kernighan (CLK)\cite{applegate2003chained}, Concorde, Lin-Kernighan-Helsgaun (LKH) \cite{helsgaun2000effective}, Multiagent Optimization System (MAOS), VSR-LKH \cite{zheng2021vsrlkhcombining},the tilted-relaxation variant of LKH (LKH-3.0.8\_0.1)\cite{ammann2026lkh-relaspeed},which we refer to as LKH-Relaxation throughout this paper, GA-EAX-restart\footnote{\url{https://github.com/senshineL/GA-EAX-restart}}. For LKH, we use the latest version:LKH-3.0.14 (February 2026) \footnote{\url{http://webhotel4.ruc.dk/~keld/research/LKH-3/}}. We observe that CLK and GA-EAX-Restart dominate the full solver pool. Therefore, in the main experiments, we use a pool of remaining five solvers, consisting of LKH, VSR-LKH, LKH-Relaxation, Concorde, and MAOS. Solver trajectories are discretized into 500 time steps. We use a starting index of $s=50$, an observation window of $L=50$ steps, and a prediction horizon of $P=400$ steps. 

\subsection{Evaluation Metrics}

We use average rank of the selected solver by our model as the primary evaluation metric. Let $s_{i,t}\in\mathcal{A}$ denote the algorithm selected by the AAAS model for
instance $i\in\mathcal{I}$ at prediction time
$t\in\mathcal{T}_{\mathrm{pred}}$. At each instance--time-step pair, all solvers
are ranked according to their incumbent objective values, with ties assigned
their average rank. Let $r_{i,t}(s)$ denote the rank of the selected solver $s$ at
$(i,t)$. The average rank achieved by the AAAS model is then defined as
\begin{equation}
R_{\mathrm{AAAS}}
=
\frac{1}{|\mathcal{I}||\mathcal{T}_{\mathrm{pred}}|}
\sum_{i\in\mathcal{I}}
\sum_{t\in\mathcal{T}_{\mathrm{pred}}}
r_{i,t}\!\left(s_{i,t}\right),
\label{eq:average_rank}
\end{equation}
where lower values indicate better performance.\\
We additionally report selection accuracy following
\citet{huerta2022improving}.
We compare against the Single Best Solver
(SBS) and Virtual Best Solver (VBS), and report the fraction of the
SBS--VBS average-rank gap closed by the model. The closed gap metric is computed using Eq.~\eqref{eq:closed_gap} in Appendix~\ref{metric}. SBS denotes the single solver with the best overall performance across all test instances and future prediction time steps, while VBS represents the oracle performance obtained by selecting the best-performing solver independently for each instance--time pair. Statistical significance against the SBS is evaluated using the Wilcoxon signed-rank test \cite{wilcoxon1992individual,demvsar2006statistical}.

\section{Results}
Our main results can be summarized in Table~\ref{tab:test_results}. Across NATIONAL, TNM, TSPLIB and VLSI, we report the average rank across future time steps and across instances, with selection accuracy following the same formula of \citet{huerta2022improving}. The Wilcoxon signed-rank test is performed based on the paired differences in average rank across instances. We also report instance-level win/tie/loss counts against the SBS, based on the average rank over the prediction horizon for each instance. Using the closed gap metric by \citet{lindauer2017open},  our model closes 46.05\% of the SBS--VBS performance gap in terms of average rank across the entire test set. To assess the contribution of the dynamic tour features, we also retrain the model without the dynamic tour features $d_{\mathrm{prev}}$ and $d_{\mathrm{next}}$. We additionally compare against Random Forest baselines using TSP features from the R package\footnote{\url{https://github.com/jakobbossek/salesperson}}, persistence baseline \ref{app:persistence}, and temporal baselines from library by \citet{wang2026deep}. Results are shown in Table~\ref{tab:rf-baseline1} and with more details including additional temporal baseline results in Appendix~\ref{app:baselines}. 

\begin{table}[h]
\centering

\caption{Comparison of ST-GNN-AAAS with selected temporal forecasting baselines, persistence, Random Forest (RF), and standalone solvers on 267 test instances over the prediction horizon. Lower average rank and higher accuracy are better.}

\label{tab:rf-baseline1}
\footnotesize
\begin{tabular}{lcc}
\toprule
Method & Average Rank & Accuracy (\%) \\
\midrule
\textbf{ST-GNN-AAAS} & $\mathbf{2.26 \pm 0.69}$ & $\mathbf{76.59}$ \\
ST-GNN-AAAS w/o tour features & $2.52 \pm 0.82$ & 69.02 \\  

\hline
Mamba ~\cite{gu2023mamba}& $2.40 \pm 0.77$ & 73.38 \\
DLinear ~\cite{dlinear} & $2.52 \pm 0.49$ & 69.99 \\
TimeFilter~\cite{hu2025timefilter}
& $2.56 \pm 0.64$ & 67.29 \\
Persistence & $2.66 \pm 1.09$ & 66.40 \\
\hline

RF\_actual-time & $3.07 \pm 0.83$ & 55.48 \\
RF\_no-time & $3.14 \pm 0.78$ & 51.81 \\
RF\_timestep & $3.10 \pm 0.76$ & 54.28 \\
\hline
VSR-LKH (SBS) & $2.61 \pm 0.58$ & 65.49 \\
CONCORDE & $2.82 \pm 1.13$ & 60.13 \\
LKH-Relaxation & $3.10 \pm 0.72$ & 52.89 \\
LKH & $3.17 \pm 0.74$ & 51.33 \\
MAOS & $3.29 \pm 0.88$ & 40.07 \\
\bottomrule
\end{tabular}
\vspace{8pt}
\caption{Performance comparison of ST-GNN-AAAS, SBS and VBS}
\label{tab:test_results}
\centering

\renewcommand{\arraystretch}{1.25}
\setlength{\tabcolsep}{0.6pt}

\resizebox{\linewidth}{!}{%
\begin{tabular}{@{}lrccccccccc ccc@{}}
\toprule
\textbf{Source} &
\textbf{$n$} &
\multicolumn{4}{c}{\textbf{Average Rank}} &
\multicolumn{4}{c}{\textbf{Accuracy (\%)}} &
\textbf{$p$-value} &
\multicolumn{3}{c}{\textbf{Model vs. SBS (\%)}} \\

\cmidrule(lr){3-6}
\cmidrule(lr){7-10}
\cmidrule(lr){12-14}

& &
\textbf{Model} &
\textbf{SBS} &
\textbf{VBS} &
\textbf{Closed (\%)} &

\textbf{Model} &
\textbf{SBS} &
\textbf{VBS} &
\textbf{Closed}&
&
\textbf{Win} &
\textbf{Tie} &
\textbf{Lose} \\

\midrule

NAT.
& 8
& $2.51 \pm 0.49$
& $3.01 \pm 0.52$
& $1.72 \pm 0.75$
&$38.76$
& $60.62 \pm 30.49$
& $46.19 \pm 41.21$
& 100
& 26.82
& $4.69\times10^{-2}$
& 62.5
& 25.0
& 12.5 \\[1pt]

TNM
& 126
& $1.81 \pm 0.65$
& $2.39 \pm 0.63$
& $1.39 \pm 0.60$ 
& $58.00$
& $69.95 \pm 32.01$
& $52.22 \pm 30.97$
& 100
& 37.11
& $7.14\times10^{-11}$
& 69.0
& 10.3
& 20.6 \\[1pt]

TSPLIB
& 70
& $2.79 \pm 0.36$
& $2.96 \pm 0.43$
& $2.58 \pm 0.62$
& $44.74$
& $90.59 \pm 19.15$
& $85.28 \pm 28.58$
& 100
& 36.07
& $1.32\times10^{-2}$
& 25.7
& 64.3
& 10.0 \\[1pt]

VLSI
& 63
& $2.52 \pm 0.44$
& $2.62 \pm 0.39$
& $1.98 \pm 0.62$
& $15.63$
& $76.35 \pm 27.18$
& $72.50 \pm 24.60$
& 100
& 14.00
& $2.71\times10^{-2}$
& 50.8
& 20.6
& 28.6 \\

\midrule

ALL
& 267
& $\mathbf{2.26 \pm 0.69}$
& $2.61 \pm 0.58$
& $1.85 \pm 0.79$
& $\mathbf{46.05}$
& $\mathbf{76.59 \pm 29.22}$
& $65.49 \pm 32.52$
& 100
& 32.16
& $6.13\times10^{-13}$
& $\mathbf{53.2}$
& 27.3
& 19.5 \\[1pt]

\bottomrule
\end{tabular}%
}
\end{table}

\bibliographystyle{plainnat}
\bibliography{references}

@article{huerta2022improving,
  author    = {Huerta, I. I. and Neira, D. A. and Ortega, D. A. and Varas, V. and Godoy, J. and Asin-Acha, R.},
  title     = {Improving the state-of-the-art in the traveling salesman problem: An anytime automatic algorithm selection},
  journal   = {Expert Systems with Applications},
  volume    = {187},
  pages     = {115948},
  year      = {2022},
  publisher = {Elsevier}
}

@article{applegate2003chained,
  author    = {Applegate, D. and Cook, W. and Rohe, A.},
  title     = {Chained Lin-Kernighan for large traveling salesman problems},
  journal   = {INFORMS Journal on Computing},
  volume    = {15},
  number    = {1},
  pages     = {82-92},
  year      = {2003},
  publisher = {INFORMS}
}

@article{helsgaun2000effective,
  author    = {Helsgaun, K.},
  title     = {An effective implementation of the Lin--Kernighan traveling salesman heuristic},
  journal   = {European Journal of Operational Research},
  volume    = {126},
  number    = {1},
  pages     = {106-130},
  year      = {2000},
  publisher = {Elsevier}
}

@article{nagata2013powerful,
  author    = {Nagata, Y. and Kobayashi, S.},
  title     = {A powerful genetic algorithm using edge assembly crossover for the traveling salesman problem},
  journal   = {INFORMS Journal on Computing},
  volume    = {25},
  number    = {2},
  pages     = {346-363},
  year      = {2013},
  publisher = {INFORMS}
}

@article{xie2008multiagent,
  author    = {Xie, X. F. and Liu, J.},
  title     = {Multiagent optimization system for solving the traveling salesman problem},
  journal   = {IEEE Transactions on Systems, Man, and Cybernetics, Part B (Cybernetics)},
  volume    = {39},
  number    = {2},
  pages     = {489-502},
  year      = {2008},
  publisher = {IEEE}
}

@article{kerschke2019automated,
  author    = {Kerschke, P. and Hoos, H. H. and Neumann, F. and Trautmann, H.},
  title     = {Automated algorithm selection: Survey and perspectives},
  journal   = {Evolutionary Computation},
  volume    = {27},
  number    = {1},
  pages     = {3-45},
  year      = {2019},
  publisher = {MIT Press}
}

@article{bai2021a3t,
  title={A3t-gcn: Attention temporal graph convolutional network for traffic forecasting},
  author={Bai, Jiandong and Zhu, Jiawei and Song, Yujiao and Zhao, Ling and Hou, Zhixiang and Du, Ronghua and Li, Haifeng},
  journal={ISPRS International Journal of Geo-Information},
  volume={10},
  number={7},
  pages={485},
  year={2021},
  publisher={MDPI}
}

@inproceedings{rozemberczki2021pytorch,
  title={PyTorch Geometric Temporal: Spatiotemporal Signal Processing with Neural Machine Learning Models},
  author={Rozemberczki, Benedek and Scherer, Paul and He, Yixuan and Panagopoulos, George and Riedel, Alexander and Astefanoaei, Maria and Kiss, Oliver and Ferenczi, Ferenc and Huk, Peter and Sarkar, Rik},
  booktitle={Proceedings of the 30th ACM International Conference on Information \& Knowledge Management},
  pages={4564--4573},
  year={2021},
  month={October},
  organization={ACM},
  doi={10.1145/3459637.3482010}
}

@inproceedings{zheng2021vsrlkhcombining,
  author = {Zheng, J. and He, K. and Zhou, J. and Jin, Y. and Li, C. M.},
  title = {Combining reinforcement learning with Lin-Kernighan-Helsgaun algorithm for the traveling salesman problem},
  booktitle = {Proceedings of the AAAI Conference on Artificial Intelligence},
  year = {2021},
  volume = {35},
  number = {14},
  pages = {12445--12452}
}

@article{ammann2026lkh-relaspeed,
  author = {Ammann, S. C. and Ostermann, B. and Stiller, S. and de Wolff, T.},
  title = {A speed-up for Helsgaun's TSP heuristic by relaxing the positive gain criterion},
  journal = {Computers \& Operations Research},
  year = {2026},
  volume = {107443}
}

@inproceedings{kotthoff2015improving,
  title={Improving the state of the art in inexact TSP solving using per-instance algorithm selection},
  author={Kotthoff, Lars and Kerschke, Pascal and Hoos, Holger and Trautmann, Heike},
  booktitle={International Conference on Learning and Intelligent Optimization},
  pages={202--217},
  year={2015},
  organization={Springer}
}

@incollection{rice1976algorithm,
  author = {Rice, J. R.},
  title = {The algorithm selection problem},
  booktitle = {Advances in Computers},
  year = {1976},
  volume = {15},
  pages = {65--118},
  publisher = {Elsevier}
}

@article{Huerta2020knapsack,
  title={Anytime automatic algorithm selection for knapsack},
  author={Huerta, Isa{\'\i}as I and Neira, Daniel A and Ortega, Daniel A and Varas, Vicente and Godoy, Julio and Asin-Acha, Roberto},
  journal={Expert Systems with Applications},
  volume={158},
  pages={113613},
  year={2020},
  publisher={Elsevier}
}

@article{Zapata_Godoy_Asin-Acha_2024,
  title={Anytime Automatic Algorithm Selection for the Multi-Agent Path Finding Problem},
  author={Zapata, Angelo and Godoy, Julio and Asin-Acha, Roberto},
  journal={IEEE Access},
  volume={12},
  pages={62177--62188},
  year={2024},
  publisher={IEEE}
}

@article{li2017diffusion,
  title={Diffusion convolutional recurrent neural network: Data-driven traffic forecasting},
  author={Li, Yaguang and Yu, Rose and Shahabi, Cyrus and Liu, Yan},
  journal={arXiv preprint arXiv:1707.01926},
  year={2017}
}

@article{demvsar2006statistical,
  title={Statistical comparisons of classifiers over multiple data sets},
  author={Dem{\v{s}}ar, Janez},
  journal={Journal of Machine learning research},
  volume={7},
  number={Jan},
  pages={1--30},
  year={2006}
}

@incollection{wilcoxon1992individual,
  title={Individual comparisons by ranking methods},
  author={Wilcoxon, Frank},
  booktitle={Breakthroughs in statistics: Methodology and distribution},
  pages={196--202},
  year={1992},
  publisher={Springer}
}

@inproceedings{lindauer2017open,
  title={Open algorithm selection challenge 2017: Setup and scenarios},
  author={Lindauer, Marius and van Rijn, Jan N and Kotthoff, Lars},
  booktitle={Open Algorithm Selection Challenge 2017},
  pages={1--7},
  year={2017},
  organization={PMLR}
}

@article{fey2019fast,
  title={Fast graph representation learning with PyTorch Geometric},
  author={Fey, Matthias and Lenssen, Jan Eric},
  journal={arXiv preprint arXiv:1903.02428},
  year={2019}
}

@article{pezo2025automatic,
  title={Automatic algorithm selection for Pseudo-Boolean optimization with given computational time limits},
  author={Pezo, Catalina and Hochbaum, Dorit and Godoy, Julio and Asin-Acha, Roberto},
  journal={Computers \& Operations Research},
  volume={173},
  pages={106836},
  year={2025},
  publisher={Elsevier}
}

@article{baldacci2012new,
  title={New state-space relaxations for solving the traveling salesman problem with time windows},
  author={Baldacci, Roberto and Mingozzi, Aristide and Roberti, Roberto},
  journal={INFORMS Journal on Computing},
  volume={24},
  number={3},
  pages={356--371},
  year={2012},
  publisher={INFORMS}
}

@incollection{applegate2011traveling,
  title={The traveling salesman problem: a computational study},
  author={Applegate, David L and Bixby, Robert E and Chv{\'a}tal, Va{\v{s}}ek and Cook, William J},
  booktitle={The traveling salesman problem},
  year={2011},
  publisher={Princeton university press}
}

@inproceedings{heins2021potential,
  title={On the potential of normalized tsp features for automated algorithm selection},
  author={Heins, Jonathan and Bossek, Jakob and Pohl, Janina and Seiler, Moritz and Trautmann, Heike and Kerschke, Pascal},
  booktitle={Proceedings of the 16th ACM/SIGEVO Conference on Foundations of Genetic Algorithms},
  pages={1--15},
  year={2021}
}

@article{mersmann2013novel,
  title={A novel feature-based approach to characterize algorithm performance for the traveling salesperson problem},
  author={Mersmann, Olaf and Bischl, Bernd and Trautmann, Heike and Wagner, Markus and Bossek, Jakob and Neumann, Frank},
  journal={Annals of Mathematics and Artificial Intelligence},
  volume={69},
  number={2},
  pages={151--182},
  year={2013},
  publisher={Springer}
}

@inproceedings{pihera2014application,
  title={Application of machine learning to algorithm selection for TSP},
  author={Pihera, Josef and Musliu, Nysret},
  booktitle={2014 IEEE 26th international conference on tools with artificial intelligence},
  pages={47--54},
  year={2014},
  organization={IEEE}
}

@inproceedings{wang2018lambdaloss,
  title={The lambdaloss framework for ranking metric optimization},
  author={Wang, Xuanhui and Li, Cheng and Golbandi, Nadav and Bendersky, Michael and Najork, Marc},
  booktitle={Proceedings of the 27th ACM international conference on information and knowledge management},
  pages={1313--1322},
  year={2018}
}

@article{burges2010ranknet,
  title={From ranknet to lambdarank to lambdamart: An overview},
  author={Burges, Christopher JC},
  journal={Learning},
  volume={11},
  number={23-581},
  pages={81},
  year={2010}
}

@article{wang2026deep,
  title={Deep time series models: A comprehensive survey and benchmark},
  author={Wang, Yuxuan and Wu, Haixu and Dong, Jiaxiang and Liu, Yong and Wang, Chen and Long, Mingsheng and Wang, Jianmin},
  journal={IEEE Transactions on Pattern Analysis and Machine Intelligence},
  year={2026},
  publisher={IEEE}
}

@article{gu2023mamba,
  title={Mamba: Linear-time sequence modeling with selective state spaces},
  author={Gu, Albert and Dao, Tri},
  journal={arXiv preprint arXiv:2312.00752},
  year={2023}
}

@article{lightts,
  title={Less is more: Fast multivariate time series forecasting with light sampling-oriented mlp structures},
  author={Zhang, Tianping and Zhang, Yizhuo and Cao, Wei and Bian, Jiang and Yi, Xiaohan and Zheng, Shun and Li, Jian},
  journal={arXiv preprint arXiv:2207.01186},
  year={2022}
}

@inproceedings{dlinear,
  title={Are transformers effective for time series forecasting?},
  author={Zeng, Ailing and Chen, Muxi and Zhang, Lei and Xu, Qiang},
  booktitle={Proceedings of the AAAI conference on artificial intelligence},
  volume={37},
  number={9},
  pages={11121--11128},
  year={2023}
}

@article{liu2023koopa,
  title={Koopa: Learning non-stationary time series dynamics with koopman predictors},
  author={Liu, Yong and Li, Chenyu and Wang, Jianmin and Long, Mingsheng},
  journal={Advances in neural information processing systems},
  volume={36},
  pages={12271--12290},
  year={2023}
}

@inproceedings{zhou2021informer,
  title={Informer: Beyond efficient transformer for long sequence time-series forecasting},
  author={Zhou, Haoyi and Zhang, Shanghang and Peng, Jieqi and Zhang, Shuai and Li, Jianxin and Xiong, Hui and Zhang, Wancai},
  booktitle={Proceedings of the AAAI conference on artificial intelligence},
  volume={35},
  number={12},
  pages={11106--11115},
  year={2021}
}

@article{hu2025timefilter,
  title={Timefilter: Patch-specific spatial-temporal graph filtration for time series forecasting},
  author={Hu, Yifan and Zhang, Guibin and Liu, Peiyuan and Lan, Disen and Li, Naiqi and Cheng, Dawei and Dai, Tao and Xia, Shu-Tao and Pan, Shirui},
  journal={arXiv preprint arXiv:2501.13041},
  year={2025}
}

@article{jung2026mambasl,
  title={MambaSL: Exploring single-layer Mamba for time series classification},
  author={Jung, Yoo-Min and Kim, Leekyung},
  journal={arXiv preprint arXiv:2604.15174},
  year={2026}
}

@article{multipatchformer,
  title={A multiscale model for multivariate time series forecasting},
  author={Naghashi, Vahid and Boukadoum, Mounir and Diallo, Abdoulaye Banire},
  journal={Scientific Reports},
  volume={15},
  number={1},
  pages={1565},
  year={2025},
  publisher={Nature Publishing Group UK, London}
}

@article{chen2023tsmixer,
  title={Tsmixer: An all-mlp architecture for time series forecasting},
  author={Chen, Si-An and Li, Chun-Liang and Yoder, Nate and Arik, Sercan O and Pfister, Tomas},
  journal={arXiv preprint arXiv:2303.06053},
  year={2023}
}

@article{vaswani2017attention,
  title={Attention is all you need},
  author={Vaswani, Ashish},
  journal={arXiv preprint arXiv:1706.03762},
  year={2017}
}

@article{patchtst,
  title={A time series is worth 64 words: Long-term forecasting with transformers},
  author={Nie, Yuqi and Nguyen, Nam H and Sinthong, Phanwadee and Kalagnanam, Jayant},
  journal={arXiv preprint arXiv:2211.14730},
  year={2022}
}

@article{pattn,
  title={Are language models actually useful for time series forecasting?},
  author={Tan, Mingtian and Merrill, Mike A and Gupta, Vinayak and Althoff, Tim and Hartvigsen, Thomas},
  journal={Advances in Neural Information Processing Systems},
  volume={37},
  pages={60162--60191},
  year={2024}
}

@article{zhou2022film,
  title={Film: Frequency improved legendre memory model for long-term time series forecasting},
  author={Zhou, Tian and Ma, Ziqing and Wen, Qingsong and Sun, Liang and Yao, Tao and Yin, Wotao and Jin, Rong and others},
  journal={Advances in neural information processing systems},
  volume={35},
  pages={12677--12690},
  year={2022}
}

@inproceedings{pyraformer,
  title={Pyraformer: Low-complexity pyramidal attention for long-range time series modeling and forecasting},
  author={Liu, Shizhan and Yu, Hang and Liao, Cong and Li, Jianguo and Lin, Weiyao and Liu, Alex X and Dustdar, Schahram},
  booktitle={International conference on learning representations},
  year={2021}
}

@article{tide,
  title={Long-term forecasting with tide: Time-series dense encoder},
  author={Das, Abhimanyu and Kong, Weihao and Leach, Andrew and Mathur, Shaan and Sen, Rajat and Yu, Rose},
  journal={arXiv preprint arXiv:2304.08424},
  year={2023}
}

@article{frets,
  title={Frequency-domain MLPs are more effective learners in time series forecasting},
  author={Yi, Kun and Zhang, Qi and Fan, Wei and Wang, Shoujin and Wang, Pengyang and He, Hui and An, Ning and Lian, Defu and Cao, Longbing and Niu, Zhendong},
  journal={Advances in neural information processing systems},
  volume={36},
  pages={76656--76679},
  year={2023}
}

@inproceedings{liu2024itransformer,
  title={itransformer: Inverted transformers are effective for time series forecasting},
  author={Liu, Yong and Hu, Tengge and Zhang, Haoran and Wu, Haixu and Wang, Shiyu and Ma, Lintao and Long, Mingsheng},
  booktitle={International conference on learning representations},
  volume={2024},
  pages={11116--11140},
  year={2024}
}

@inproceedings{wang2024timemixer,
  title={Timemixer: Decomposable multiscale mixing for time series forecasting},
  author={Wang, Shiyu and Wu, Haixu and Shi, Xiaoming and Hu, Tengge and Luo, Huakun and Ma, Lintao and Zhang, James and Zhou, Jun},
  booktitle={International conference on learning representations},
  volume={2024},
  pages={38626--38652},
  year={2024}
}

@article{wang2024timexer,
  title={Timexer: Empowering transformers for time series forecasting with exogenous variables},
  author={Wang, Yuxuan and Wu, Haixu and Dong, Jiaxiang and Qin, Guo and Zhang, Haoran and Liu, Yong and Qiu, Yunzhong and Wang, Jianmin and Long, Mingsheng},
  journal={Advances in neural information processing systems},
  volume={37},
  pages={469--498},
  year={2024}
}

@inproceedings{wang2023micn,
  title={Micn: Multi-scale local and global context modeling for long-term series forecasting},
  author={Wang, Huiqiang and Peng, Jian and Huang, Feihu and Wang, Jince and Chen, Junhui and Xiao, Yifei},
  booktitle={The eleventh international conference on learning representations},
  year={2023}
}

@article{segrnn,
  title={Segrnn: Segment recurrent neural network for long-term time series forecasting},
  author={Lin, Shengsheng and Lin, Weiwei and Wu, Wentai and Zhao, Feiyu and Mo, Ruichao and Zhang, Haotong},
  journal={IEEE Internet of Things Journal},
  year={2025},
  publisher={IEEE}
}

@inproceedings{zhou2022fedformer,
  title={Fedformer: Frequency enhanced decomposed transformer for long-term series forecasting},
  author={Zhou, Tian and Ma, Ziqing and Wen, Qingsong and Wang, Xue and Sun, Liang and Jin, Rong},
  booktitle={International conference on machine learning},
  pages={27268--27286},
  year={2022},
  organization={PMLR}
}

@article{tft,
  title={Temporal fusion transformers for interpretable multi-horizon time series forecasting},
  author={Lim, Bryan and Ar{\i}k, Sercan {\"O} and Loeff, Nicolas and Pfister, Tomas},
  journal={International journal of forecasting},
  volume={37},
  number={4},
  pages={1748--1764},
  year={2021},
  publisher={Elsevier}
}

@article{liu2022scinet,
  title={Scinet: Time series modeling and forecasting with sample convolution and interaction},
  author={Liu, Minhao and Zeng, Ailing and Chen, Muxi and Xu, Zhijian and Lai, Qiuxia and Ma, Lingna and Xu, Qiang},
  journal={Advances in neural information processing systems},
  volume={35},
  pages={5816--5828},
  year={2022}
}

@article{non-stationary,
  title={Non-stationary transformers: Exploring the stationarity in time series forecasting},
  author={Liu, Yong and Wu, Haixu and Wang, Jianmin and Long, Mingsheng},
  journal={Advances in neural information processing systems},
  volume={35},
  pages={9881--9893},
  year={2022}
}

@inproceedings{cai2024msgnet,
  title={Msgnet: Learning multi-scale inter-series correlations for multivariate time series forecasting},
  author={Cai, Wanlin and Liang, Yuxuan and Liu, Xianggen and Feng, Jianshuai and Wu, Yuankai},
  booktitle={Proceedings of the AAAI conference on artificial intelligence},
  volume={38},
  number={10},
  pages={11141--11149},
  year={2024}
}

@inproceedings{zhang2023crossformer,
  title={Crossformer: Transformer utilizing cross-dimension dependency for multivariate time series forecasting},
  author={Zhang, Yunhao and Yan, Junchi},
  booktitle={The eleventh international conference on learning representations},
  year={2023}
}

@article{timesnet,
  title={Timesnet: Temporal 2d-variation modeling for general time series analysis},
  author={Wu, Haixu and Hu, Tengge and Liu, Yong and Zhou, Hang and Wang, Jianmin and Long, Mingsheng},
  journal={arXiv preprint arXiv:2210.02186},
  year={2022}
}

@article{wu2021autoformer,
  title={Autoformer: Decomposition transformers with auto-correlation for long-term series forecasting},
  author={Wu, Haixu and Xu, Jiehui and Wang, Jianmin and Long, Mingsheng},
  journal={Advances in neural information processing systems},
  volume={34},
  pages={22419--22430},
  year={2021}
}


\appendix

\section{Appendix}

\subsection{Dataset and Solver Execution Runs}
\label{dataset}
The dataset of our work comes from Huerta et al. \cite{huerta2022improving}. 
To be specific, the TSP instances include four datasets: RUE (Random Uniform Euclidean), NETGEN, NETGENM and TSPGEN.
The test instances come from public benchmark libraries including TSPLIB\footnote{\url{http://comopt.ifi.uni-heidelberg.de/software/TSPLIB95/tsp/}}, 
TNM\footnote{\url{https://www.or.uni-bonn.de/~hougardy/HardTSPInstances.html}}, NATIONAL\footnote{\url{https://www.math.uwaterloo.ca/tsp/world/countries.html}} and VLSI\footnote{\url{http://www.math.uwaterloo.ca/tsp/vlsi/index.html}}. 
In addition to the TSP instances, iterative solution trajectories, only including the fitness values of the new best solutions, achieved by a set of TSP solvers are provided.
However, several inconsistencies have been identified in those public records. 
For example, most MAOS runs on the NATIONAL datasets are unavailable with the corresponding files showing Java execution failures. 
Furthermore, some TSPLIB execution files show inconsistent objective values across solvers. For instance, in MAOS/TSPLIB/fl417.tsp.out \footnote{\url{https://isaiash.github.io/anytime_tsp/}}, MAOS reports a tour cost of 2085, while CLK, Concorde, EAX, and LKH all converge to a cost around 11861. 
However, the optimal value\footnote{\url{https://comopt.ifi.uni-heidelberg.de/software/TSPLIB95/tsp/TSP-BEST.html}} of fl417.tsp is reported to be 11861.
We therefore generate those time-new best fitness values data on those TSP instances, capped at 5000 cities, using with updated TSP solver portfolio.
Beyond solution qualities, actual TSP solution permutations are maintained as they are required in the proposed Spatio-Temporal Graph Neural Network (ST-GNN) architecture setup.
The complete details are provided in Table~\ref{tab:benchmark}.

\begin{table}[h]
\caption{Benchmark composition and retained TSP instances.}
\label{tab:benchmark}
\centering

\renewcommand{\arraystretch}{1.20}
\setlength{\tabcolsep}{4pt}

\small
\begin{tabular}{lrrrrcr}
\toprule
\textbf{Source} &
\multicolumn{4}{c}{\textbf{Number of instances}} &
\multicolumn{2}{c}{\textbf{Retained set}} \\

\cmidrule(lr){2-5}
\cmidrule(lr){6-7}

&
\textbf{Original} &
\textbf{$n\leq5000$} &
\textbf{\textsc{euc\_2d}} &
\textbf{Retained} &
\textbf{$n$ range} &
\textbf{Median $n$} \\

\midrule

\multicolumn{7}{l}{\textit{Training set}} \\

NETGEN
& 600
& 600
& 600
& 600
& 500--2{,}000
& 1{,}250 \\

NETGENM
& 600
& 600
& 600
& 600
& 500--2{,}000
& 1{,}250 \\

RUE
& 3150
& 3150
& 3150
& 3150
& 500--2{,}000
& 1{,}500 \\

TSPGEN
& 1981
& 499
& 499
& 496
& 31--4{,}999
& 2{,}518 \\

\textbf{Training subtotal}
& \textbf{6331}
& \textbf{4849}
& \textbf{4849}
& \textbf{4846}
& --
& -- \\

\midrule

\multicolumn{7}{l}{\textit{Test set }} \\

NATIONAL
& 27
& 10
& 10
& 8
& 194--4{,}663
& 1{,}300 \\

TNM
& 141
& 126
& 126
& 126
& 52--4{,}000
& 239 \\

TSPLIB
& 88
& 78
& 71
& 70
& 51--4{,}461
& 308 \\

VLSI
& 102
& 63
& 63
& 63
& 131--4{,}966
& 2{,}217 \\

\textbf{Test subtotal}
& \textbf{358}
& \textbf{277}
& \textbf{270}
& \textbf{267}
& --
& -- \\

\midrule

\textbf{Total}
& \textbf{6689}
& \textbf{5126}
& \textbf{5119}
& \textbf{5113}
& --
& -- \\

\bottomrule
\end{tabular}%

\end{table}

Following the similar solver execution setup as \citet{huerta2022improving}, we ran each instance under the same outer control condition. The conditions are defined as: the solver is allowed to run at most 7200 seconds for one each instance and if no better objective value is found in ten minutes, the execution for this instance will be terminated. We make only minimal modifications to the original solver implementations. Specifically, we add logging to record the objective value, execution time, and tour permutation of each incumbent solution. We also increase the relevant outer-loop limits so that the solvers do not terminate prematurely under their default settings before potentially finding better solutions. The modified parameters are summarized in Table~\ref{tab:solver-params}. We made our dataset of TSP execution runs available at \url{https://huggingface.co/datasets/anonymous-stgnn-aas/TSP_EXECUTION_RUNS/tree/main}
\begin{table}[!t]
\caption{Solver Parameters Modified to Prevent Premature Termination}
\label{tab:solver-params}
\centering
\renewcommand{\arraystretch}{1.15}
\small
\begin{tabular}{@{}llrr@{}}
\toprule
Solver & Parameter & Value & Default \\
\midrule
LKH, VSR-LKH, LKH-Rlx. & \texttt{RUNS} & 99999999 & 10 \\
CLK (\texttt{linkern})  & \texttt{-R}   & 99999999 & $n$ \\
MAOS                    & \texttt{T}    & 99999999 & 500 \\
\bottomrule
\end{tabular}

\vspace{2pt}
\begin{minipage}{\columnwidth}
\footnotesize
\textit{Note:} Each bound is raised beyond what is reachable within the time
budget (\texttt{RUNS}: restarts; \texttt{-R}: LK kicks; \texttt{T}: MAOS
cycles), so the effective stopping rule is the external budget---a 2\,h
wall-clock cap and a 600\,s no-improvement cutoff enforced by the runner---not
a solver-internal counter. All other settings follow the default values.
\end{minipage}
\end{table}

\subsubsection{Solver Dynamics}
\label{solver_dynamics}
\begin{figure}[htbp]
    \centering

    \includegraphics[width=0.78\linewidth]{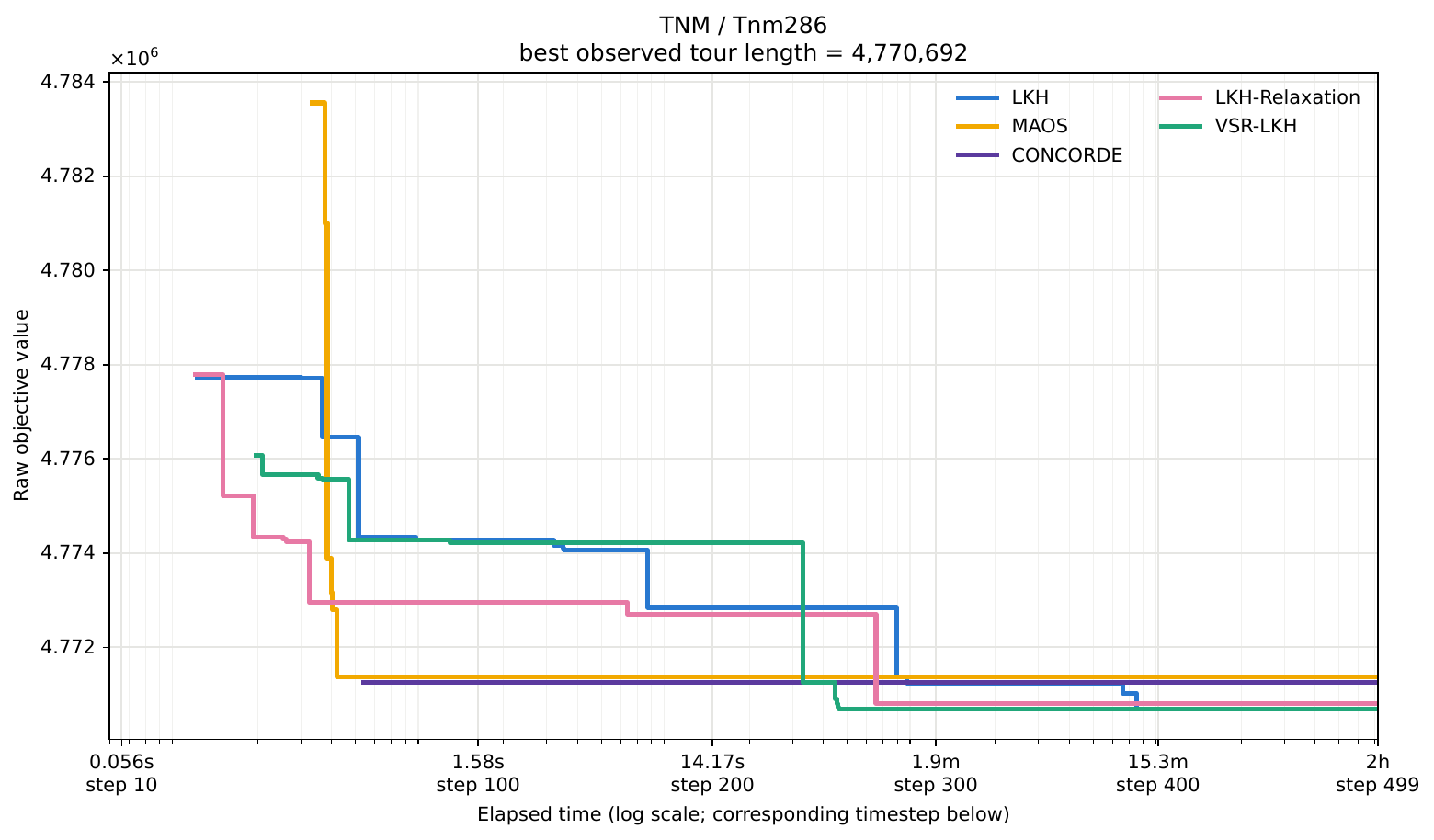}

    \vspace{2mm}

    \includegraphics[width=0.78\linewidth]{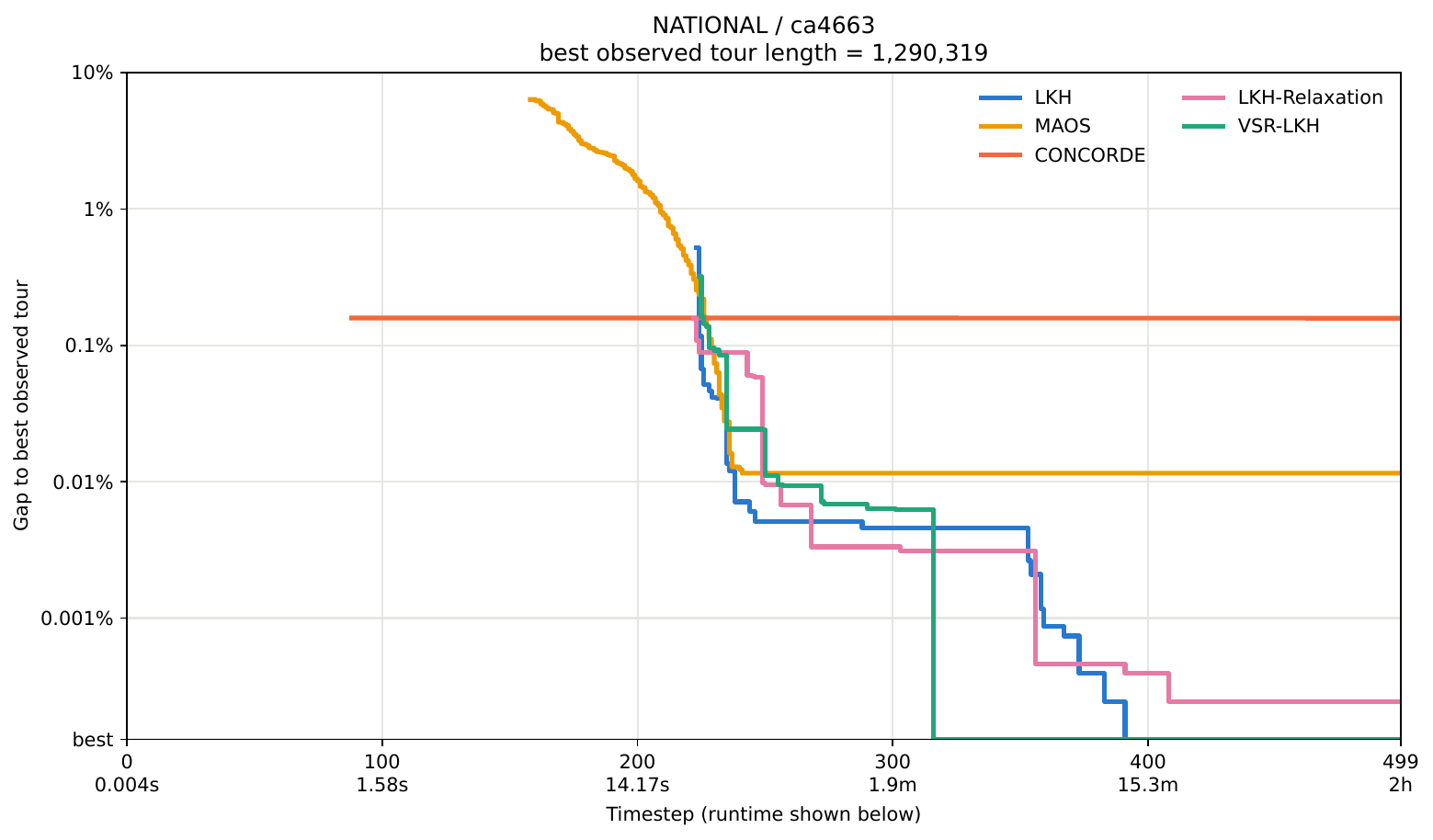}

    \caption{
    Temporal behavior of the five selected TSP solvers on particular instances.
    Top: raw solution cost on TNM286.
    Bottom: optimality gap on the NATIONAL instance ca4663.
    }
    \label{fig:solver_dynamics}
\end{figure}

\begin{figure}[t]
    \centering

    \includegraphics[width=\linewidth]{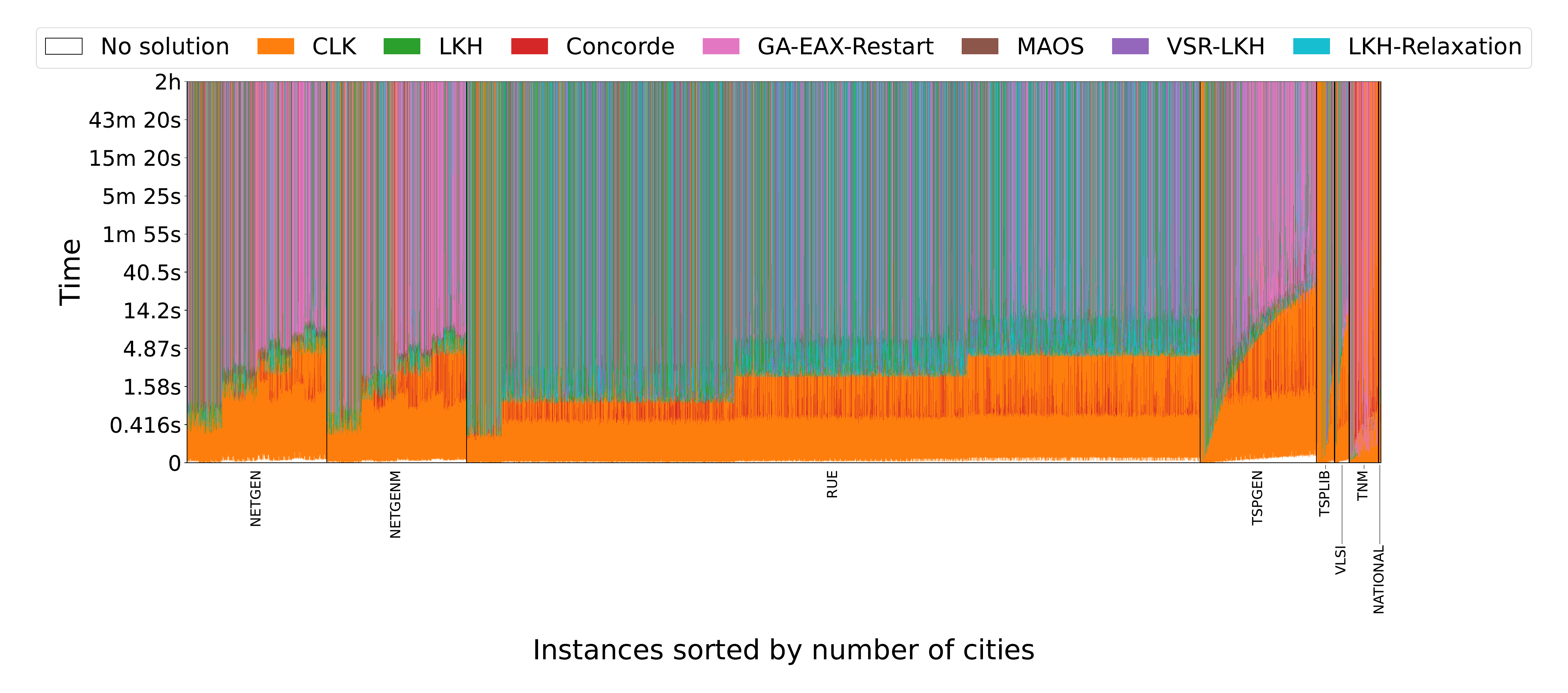}


    \includegraphics[width=\linewidth]{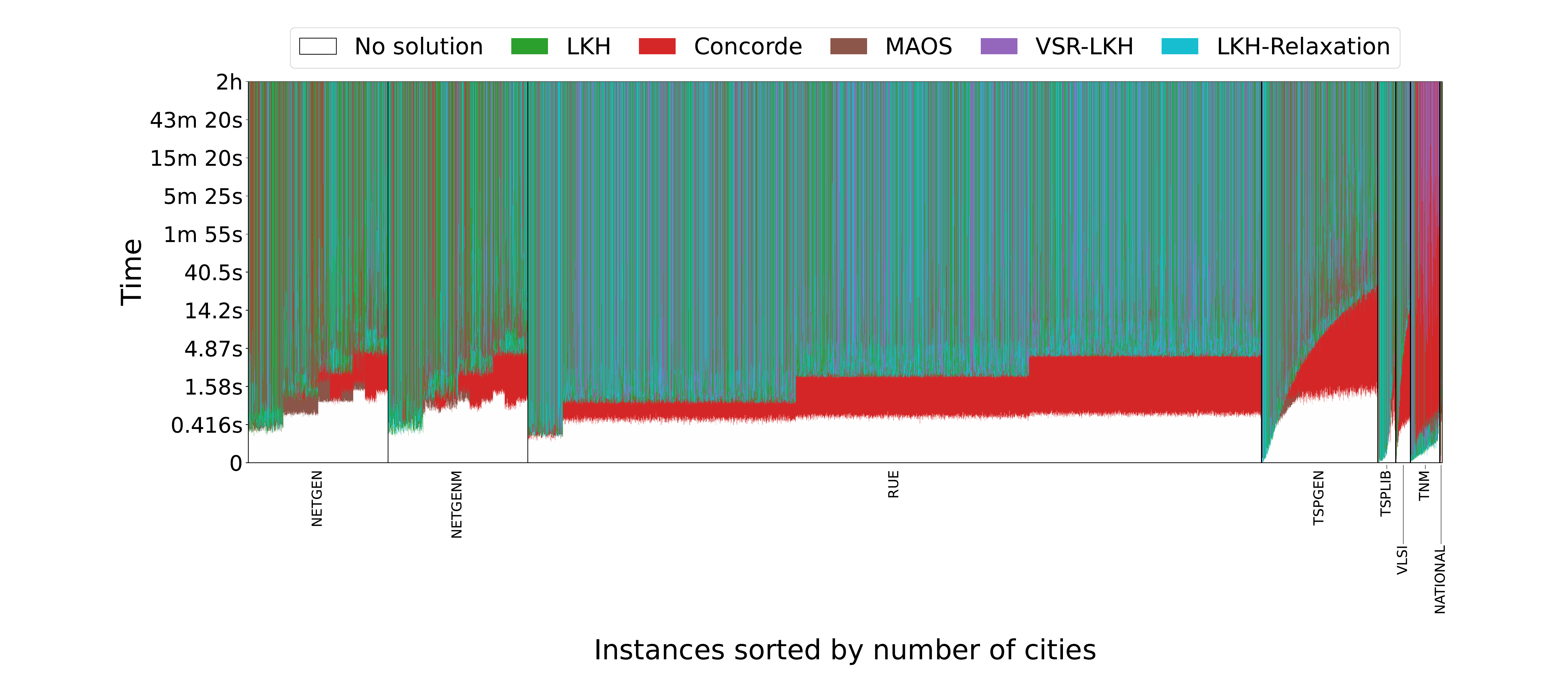}

    \caption{
    Temporal dynamics of the leading solver across selected instances with different solver pools.
    The upper and lower panels correspond to the seven-solver and five-solver portfolios, respectively.
    Colors indicate the solver achieving the best incumbent solution at each time point, while the
    x-axis is ordered by instance size within each dataset.
    }
    \label{fig:solver_pool_dynamics}
\end{figure}

To illustrate solver dynamics over time, we visualize the temporal behavior of different solver portfolios as well as several representative instances. See Figure~\ref{fig:solver_dynamics}
and Figure~\ref{fig:solver_pool_dynamics}




\begin{table}[!htb]
\centering
\caption{Performance of fixed solvers and the VBS over the prediction horizon
}
\label{training_testing_cal}
\small
\setlength{\tabcolsep}{5pt}
\renewcommand{\arraystretch}{1.05}
\begin{tabular}{l | cc | cc}
\toprule
Solver & \multicolumn{2}{c}{Training set ($n=4846$)} & \multicolumn{2}{c}{Test set ($n=267$)} \\
\cmidrule(lr){2-3} \cmidrule(lr){4-5}
                        & Avg Rank & Accuracy (\%) & Avg Rank & Accuracy (\%) \\
\midrule
LKH             & \textbf{2.48} & \textbf{77.11} & 3.17 & 51.33 \\
LKH-Relaxation  & 2.51 & 76.07 & 3.10 & 52.89 \\
VSR-LKH         & 2.70 & 70.99 & \textbf{2.61} & \textbf{65.49} \\
CONCORDE        & 3.62 & 48.25 & 2.82 & 60.13 \\
MAOS            & 3.69 & 26.93 & 3.29 & 40.07 \\
\midrule
VBS (Oracle)           & 2.00 & 100.00 & 1.85 & 100.00 \\
\bottomrule
\end{tabular}
\end{table}

\subsubsection{Distribution Shift}
\begin{figure}[t]
    \centering
    \includegraphics[width=0.7\linewidth]{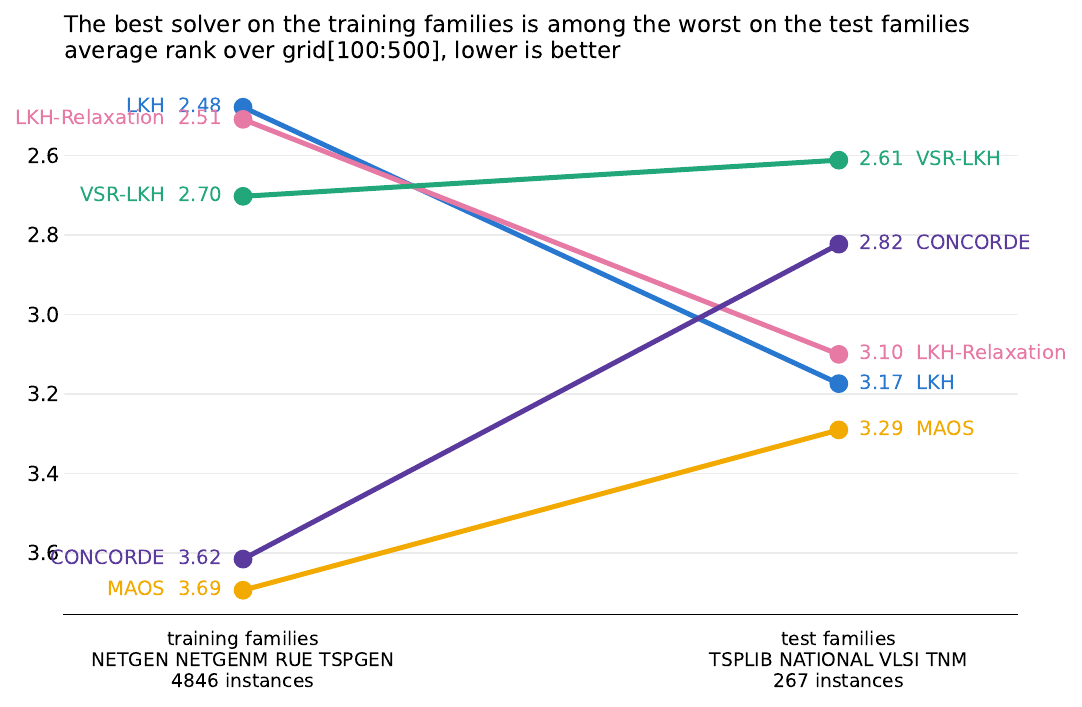}
    \caption{Rank inversion of solvers between training and test datasets.}
    \label{fig:rankinversion}
\end{figure}

We observe a possible distribution shift between the training and testing datasets. Table~\ref{training_testing_cal} compares the average rank and accuracy of each fixed solver on the training and test sets over the same prediction window. Figure~\ref{fig:rankinversion} further visualizes this shift.

\subsection{Training Details}
 Experiments including solver execution across instances were performed on a shared server equipped with two AMD EPYC 7763 processors (64 cores each, 2.45 GHz base / 3.50 GHz maximum boost, 2 NUMA nodes) and 314 GiB of RAM, running Ubuntu 22.04.5 LTS (kernel 5.15.0, x86-64; GCC 11.4.0, OpenJDK 20.0.2). Training and testing are conducted on the same server with a single NVIDIA GeForce RTX 3090 GPU (24 GB). 
 The backbone of spatio-temporal encoder is implementation of A3TGCN \cite{bai2021a3t} from \cite{rozemberczki2021pytorch}\footnote{\url{https://pytorch-geometric-temporal.readthedocs.io/en/}}. 
 The code is implemented in PyTorch Geometric to handle TSP instances of varying sizes \cite{fey2019fast}.

 15\% of training data are used for validation. 
 The model is trained for 100 epochs, and the checkpoint with the best validation average rank is selected for evaluation. 
 We use AdamW as the optimizer with a learning rate of $2\times10^{-4}$ and a weight decay of $1\times10^{-4}$. 
 A \texttt{CosineAnnealingLR} \footnote{\url{https://docs.pytorch.org/docs/2.13/generated/torch.optim.lr_scheduler.CosineAnnealingLR.html}} scheduler is employed to adjust the learning rate during training. Our model has a relatively small number of parameters, with 296,395 parameters in total.
 
\textbf{Missing solutions and training.}
A time step is considered valid if at least one solver has produced a feasible solution. 
For a solver without a feasible solution at a valid time step, we assign a penalized objective value equal to $1.2$ times the worst available objective value at that time step. 
Its permutation-based node features are set to zero while the static coordinates are retained. 
All the evaluation of our model is based on valid time steps and invalid time steps will be filtered. 
Under the Anytime Automatic Algorithm Selection (AAAS) framework, the evaluation metrics also incorporate more granular, time-step-level information, with Single Best Solver (SBS) evaluated across the same set of datasets and all time steps, Virtual Best Solver (VBS) as the oracle performance of choosing the best solver across all the timesteps \cite{Zapata_Godoy_Asin-Acha_2024}. We train ST-GNN-AAAS using a pairwise ranking loss. For each instance and time budget, the loss encourages higher scores for solvers with lower tour costs and weights each pair by its relative cost difference, capped at 3, inspired by \citep{wang2018lambdaloss}.

\subsection{AAAS Evaluation Metrics}
\label{metric}
\citet{huerta2022improving} first proposed their AAAS approach, also for the Traveling Salesman Problem (TSP), in order to balance the trade-off between speed and solution quality. The primary evaluation criterion is the accuracy of selecting a best-performing solver at each time step:
\begin{equation}
\mathrm{accuracy}(s)
=
\frac{
\sum_{t=1}^{500}
\mathrm{equal}\!\left(cost_s(t), best\_cost(t)\right)
}{500},
\label{eq:solver_accuracy}
\end{equation}
where $\mathrm{equal}(a,b) = \mathbf{1}_{a=b}$,
$cost_s(t)$ denotes the incumbent objective value of solver $s$ at time step $t$, and $best\_cost(t)$ denotes the best incumbent objective value achieved among all solvers at the same time step. 
Concerning $\mathrm{equal}(\cdot,\cdot)$ returning 1 when the two objective values are equal and 0 otherwise, this metric can actually be problematic as different solvers might converge to the same objective value, especially in late stages. 
Therefore, in such cases, the selection of any tied solver is considered correct, leading to high accuracy. 
Subsequent studies on AAAS have considered additional performance dimensions beyond the selection accuracy. \citet{Zapata_Godoy_Asin-Acha_2024} in his AAAS work on Multi-Agent Path Finding Problem proposes metrics to measure the performance of meta-solver on an average level
\begin{equation}
\hat{m}
=
\frac{
M_{\mathrm{Meta}} - M_{\mathrm{VBS}}
}{
M_{\mathrm{SBS}} - M_{\mathrm{VBS}}
},
\label{eq:normalized_performance}
\end{equation}
where $M_{\mathrm{Meta}}$, $M_{\mathrm{SBS}}$, and $M_{\mathrm{VBS}}$ denote the aggregated performance measures of the meta-solver, the single best solver (SBS), and the virtual best solver (VBS), respectively. Similarly, \citet{pezo2025automatic} evaluate the solution quality of the selected solver by aggregating normalized objective values over instance--time-step pairs. 

Inspired by \citet{pezo2025automatic}, we follow this idea with a slightly different formulation using our notation and restrict the evaluation to the future prediction window $T_{\mathrm{pred}}$, corresponding to time steps 100--500 in our experiments. Let $o_s(i,t)$ denote the objective value obtained by solver $s \in \mathcal{A}$ on instance $i \in \mathcal{I}$ at time $t \in T_{\mathrm{pred}}$. For each instance, the range of feasible objective values over all solvers and future prediction time steps is defined as

\begin{equation}
o_{\min}(i)
=
\min_{s \in \mathcal{A},\, t \in T_{\mathrm{pred}}}
o_s(i,t),
\qquad
o_{\max}(i)
=
\max_{s \in \mathcal{A},\, t \in T_{\mathrm{pred}}}
o_s(i,t),
\label{eq:objective_range}
\end{equation}

For the solver $s_{i,t}$ selected, its normalized objective value at $(i,t)$ is

\begin{equation}
\tilde{o}_{i,t}
=
\frac{
o_{s_{i,t}}(i,t)-o_{\min}(i)
}{
o_{\max}(i)-o_{\min}(i)
}.
\label{eq:normalized_objective}
\end{equation}

In an anytime setting, different solvers may find feasible solutions at different time steps across instances. To account for such cases, we follow \citet{pezo2025automatic} and assign a penalty value when the selected solver has not yet found a feasible solution:
\begin{equation}
q_{s_{i,t}}
=
\begin{cases}
0,
& \text{if } o_{\min}(i)=o_{\max}(i), \\[3pt]
2,
& \text{if } o_{s_{i,t}}(i,t) \text{ is undefined}, \\[6pt]
\tilde{o}_{i,t},
& \text{otherwise}.
\end{cases}
\label{eq:normalized_quality}
\end{equation}
Here, $q_{s_{i,t}}$ denotes the normalized score of solver $s$ for an instance--time-step pair $(i,t)$. We then aggregate these scores over all instances and prediction time steps. For our meta-solver, ST-GNN-AAAS, let $s'_{i,t} \in \mathcal{A}$ denote the solver selected for instance $i$ at prediction time $t$. The average normalized solution quality of STGNN-AAAS is then defined as

\begin{equation}
Q_{\mathrm{AAAS}}
=
\frac{1}{|\mathcal{I}|\,|T_{\mathrm{pred}}|}
\sum_{i \in \mathcal{I}}
\sum_{t \in T_{\mathrm{pred}}}
q_{s'_{i,t},\,i,t},
\label{eq:normalized_quality_aaas}
\end{equation}

These metrics provide a richer evaluation than selection accuracy alone because
they account for the performance loss associated with suboptimal solver choices. In this work, we additionally report average rank as a complementary, scale-independent metric that reflects the relative performance of the selected solver at each instance--time-step pair, borrowing ideas from \citet{demvsar2006statistical}. Meanwhile, the closed gap formula by \citet{lindauer2017open} is also reported correspondingly.
\begin{equation}
\mathrm{Closed Gap}
=
\frac{
m_{\mathrm{SBS}}-m_{\mathrm{Model}}
}{
m_{\mathrm{SBS}}-m_{\mathrm{VBS}}
}.
\label{eq:closed_gap}
\end{equation}

\section{Additional Baseline Comparisons}
\label{app:baselines}

\subsection{Random Forest Baselines}
\label{rf}
For evaluating the effectiveness of our ST-GNN-AAAS model, we compare against classical machine learning (ML) algorithms based on handcrafted features. 
To derive the handcrafted features of TSP instances, we use package salesperson implemented in R \footnote{\url{https://github.com/jakobbossek/salesperson}}, including 167 TSP features. 
We choose Random Forest (RF) as the classical ML baseline because it has shown strong and robust performance in previous TSP algorithm-selection studies \cite{heins2021potential,kotthoff2015improving,mersmann2013novel, pihera2014application} and has also been used in prior AAAS works \cite{pezo2025automatic,Huerta2020knapsack}.
The model input for a given (instance, time) pair contains those handcrafted TSP features  along with the specific timestep, following similar setup in \citet{pezo2025automatic}
Since ties happen often, when constructing the supervised label, ties are broken in favor of the solver that first achieves the same incumbent value. 
In this case, the label for each (instance, time) pair is determined.
To ensure a fair comparison, the split of the training dataset is the same as ST-GNN-AAAS.
\paragraph{RF\_actual-time.}
The resulting RF classifier is built with the hyper-parameter setting of
\texttt{n\_estimators=100}, \texttt{max\_features="sqrt"}, and \texttt{criterion="gini"}.

\paragraph{RF\_no-time.}
The same RF setup as \textbf{RF\_actual-time}, yet without the temporal feature.

\paragraph{RF\_timestep.}
The same RF setup as \textbf{RF\_actual-time}, with the time-step index provided as an additional temporal feature instead of the actual elapsed time.
\paragraph{Feature importance.}
For \textbf{RF\_actual-time}, the 5 most important feature groups are revealed as
time (0.235), nearest-neighbor graph (0.200), convex-hull
(0.116), minimum-spanning-tree (0.109), and angle-based features
(0.100).
These are determined using impurity-based feature importance rounded to 3 decimal digits. 
We made the extraction of features available at \url{https://huggingface.co/datasets/anonymous-stgnn-aas/TSP_EXECUTION_RUNS/tree/main/salesperson_features}
\subsection{Persistence and Temporal Baselines}

\paragraph{Persistence baseline.}
\label{app:persistence}
In addition to the classical machine-learning baselines, we compare against a simple
persistence heuristic that uses the same observation window as ST-GNN-AAAS. For each instance and at the last
observed step ($t \approx 1.54$\,s), it selects the solver with the lowest incumbent tour
cost and keeps that solver for every step of the prediction window. When several solvers
share the lowest cost, the one that reached it first is chosen. The heuristic needs no
training, and it tests whether the solver leading early in the run remains the best choice later on.

\paragraph{Temporal forecasting baselines.}
\label{temporal_baselines}
We additionally evaluate temporal forecasting models
implemented using the Time Series Library
(TSLib)~\cite{wang2026deep}.
Table~\ref{tab:five-solver-uniform-labels} reports the complete
comparison. All methods are evaluated on the same
267 test instances over the prediction horizon.

\begin{table}[H]
\centering
\caption{Solver selection on 267 test instances with a five-solver portfolio.}
\label{tab:five-solver-uniform-labels}

\begingroup
\small
\setlength{\tabcolsep}{0pt}
\renewcommand{\arraystretch}{1.08}

\newcommand{\solverlabel}[1]{\makebox[.129\linewidth][l]{#1}}
\newcommand{\sixcell}[1]{\makebox[.144\linewidth][c]{\shortstack{#1}}}
\newcommand{\fivecell}[1]{\makebox[.173\linewidth][c]{\shortstack{#1}}}

\newcommand{\solverrank}[2]{\ensuremath{#1\,{\pm}\,#2}}

\newcommand{\tsyear}[1]{%
  \begingroup
  \hypersetup{hidelinks}%
  \setlength{\fboxsep}{0.65pt}%
  \setlength{\fboxrule}{0.65pt}%
  \scriptsize
  \fcolorbox{green!100!black}{white}{\strut\citeyear{#1}}%
  \endgroup
}

\begin{tabular}{@{}l|*{6}{c}@{}}
\toprule
\multicolumn{1}{@{}l}{}
  & \multicolumn{2}{c}{\textbf{ST-GNN-AAAS(ours)}}
  & \multicolumn{4}{c}{\textbf{Persistence and random forests}} \\
\cmidrule(lr){2-3}\cmidrule(l){4-7}
\solverlabel{Models}
  & \sixcell{w/ tour\\features}
  & \sixcell{w/o tour\\features}
  & \sixcell{Persistence}
  & \sixcell{RF\\(actual time)}
  & \sixcell{RF\\(time-step)}
  & \sixcell{RF\\(no time)} \\
\solverlabel{Avg.\ rank $\downarrow$}
  & \solverrank{\textcolor{red!55!black}{\mathbf{2.26}}}{0.69}
  & \solverrank{2.52}{0.82}
  & \solverrank{2.66}{1.09}
  & \solverrank{3.07}{0.83}
  & \solverrank{3.10}{0.76}
  & \solverrank{3.14}{0.78} \\
\solverlabel{Acc.\ (\%) $\uparrow$}
  & \textcolor{red!55!black}{\textbf{76.59}} & 69.02 & 66.40 & 55.48 & 54.28 & 51.81 \\
\end{tabular}\par\nointerlineskip
%
\begin{tabular}{@{}l|*{5}{c}@{}}
\toprule
\multicolumn{6}{@{}l}{\textbf{Solver Algorithms}} \\
\midrule
\solverlabel{}
  & \fivecell{VSR-LKH\\(SBS)}
  & \fivecell{CONCORDE}
  & \fivecell{LKH-\\Relaxation}
  & \fivecell{LKH}
  & \fivecell{MAOS} \\
\solverlabel{Avg.\ rank $\downarrow$}
  & \solverrank{2.61}{0.58}
  & \solverrank{2.82}{1.13}
  & \solverrank{3.10}{0.72}
  & \solverrank{3.17}{0.74}
  & \solverrank{3.29}{0.88} \\
\solverlabel{Acc.\ (\%) $\uparrow$}
  & 65.49 & 60.13 & 52.89 & 51.33 & 40.07 \\
\end{tabular}\par\nointerlineskip
%
\begin{tabular}{@{}l|*{6}{c}@{}}
\toprule
\multicolumn{7}{@{}l}{\textbf{Temporal baselines}} \\
\midrule

\solverlabel{Models}
  & \sixcell{Mamba}
  & \sixcell{MambaSimple}
  & \sixcell{LightTS}
  & \sixcell{DLinear}
  & \sixcell{Koopa}
  & \sixcell{Informer} \\
  & \tsyear{gu2023mamba}
  & \tsyear{gu2023mamba}
  & \tsyear{lightts}
  & \tsyear{dlinear}
  & \tsyear{liu2023koopa}
  & \tsyear{zhou2021informer} \\
\solverlabel{Avg.\ rank $\downarrow$}
  & \solverrank{\textcolor{blue!55!black}{\underline{2.40}}}{0.77}
  & \solverrank{2.47}{0.64}
  & \solverrank{2.48}{0.55}
  & \solverrank{2.52}{0.49}
  & \solverrank{2.52}{0.61}
  & \solverrank{2.54}{0.44} \\
\solverlabel{Acc.\ (\%) $\uparrow$}
  & \textcolor{blue!55!black}{\underline{73.38}} & 71.56 & 69.42 & 69.99 & 69.20 & 68.80 \\
\midrule

\solverlabel{Models}
  & \sixcell{TimeFilter}
  & \sixcell{MambaSL}
  & \sixcell{MultiPatch\\Former}
  & \sixcell{TSMixer}
  & \sixcell{Transformer}
  & \sixcell{PatchTST} \\
  & \tsyear{hu2025timefilter}
  & \tsyear{jung2026mambasl}
  & \tsyear{multipatchformer}
  & \tsyear{chen2023tsmixer}
  & \tsyear{vaswani2017attention}
  & \tsyear{patchtst} \\
\solverlabel{Avg.\ rank $\downarrow$}
  & \solverrank{2.56}{0.64}
  & \solverrank{2.58}{0.54}
  & \solverrank{2.60}{0.91}
  & \solverrank{2.61}{0.45}
  & \solverrank{2.62}{0.48}
  & \solverrank{2.62}{0.80} \\
\solverlabel{Acc.\ (\%) $\uparrow$}
  & 67.29 & 67.03 & 66.30 & 67.20 & 66.35 & 66.56 \\
\midrule

\solverlabel{Models}
  & \sixcell{PAttn}
  & \sixcell{FiLM}
  & \sixcell{Pyraformer}
  & \sixcell{TiDE}
  & \sixcell{FreTS}
  & \sixcell{iTransformer} \\
  & \tsyear{pattn}
  & \tsyear{zhou2022film}
  & \tsyear{pyraformer}
  & \tsyear{tide}
  & \tsyear{frets}
  & \tsyear{liu2024itransformer} \\
\solverlabel{Avg.\ rank $\downarrow$}
  & \solverrank{2.68}{0.85}
  & \solverrank{2.69}{0.81}
  & \solverrank{2.72}{0.43}
  & \solverrank{2.74}{0.84}
  & \solverrank{2.74}{0.56}
  & \solverrank{2.76}{0.88} \\
\solverlabel{Acc.\ (\%) $\uparrow$}
  & 64.44 & 64.45 & 64.07 & 62.62 & 63.50 & 63.04 \\
\midrule

\solverlabel{Models}
  & \sixcell{TimeMixer}
  & \sixcell{TimeXer}
  & \sixcell{MICN}
  & \sixcell{SegRNN}
  & \sixcell{FEDformer}
  & \sixcell{TFT} \\
  & \tsyear{wang2024timemixer}
  & \tsyear{wang2024timexer}
  & \tsyear{wang2023micn}
  & \tsyear{segrnn}
  & \tsyear{zhou2022fedformer}
  & \tsyear{tft} \\
\solverlabel{Avg.\ rank $\downarrow$}
  & \solverrank{2.77}{0.87}
  & \solverrank{2.78}{0.91}
  & \solverrank{2.85}{0.46}
  & \solverrank{2.86}{0.55}
  & \solverrank{2.87}{0.53}
  & \solverrank{2.92}{0.76} \\
\solverlabel{Acc.\ (\%) $\uparrow$}
  & 61.19 & 59.68 & 60.95 & 61.13 & 60.42 & 55.00 \\
\midrule

\solverlabel{Models}
  & \sixcell{SCINet}
  & \sixcell{Nonstat.\\Transformer}
  & \sixcell{MSGNet}
  & \sixcell{Crossformer}
  & \sixcell{TimesNet}
  & \sixcell{Autoformer} \\
  & \tsyear{liu2022scinet}
  & \tsyear{non-stationary}
  & \tsyear{cai2024msgnet}
  & \tsyear{zhang2023crossformer}
  & \tsyear{timesnet}
  & \tsyear{wu2021autoformer} \\
\solverlabel{Avg.\ rank $\downarrow$}
  & \solverrank{2.94}{0.76}
  & \solverrank{2.95}{0.84}
  & \solverrank{3.00}{0.69}
  & \solverrank{3.05}{0.72}
  & \solverrank{3.10}{0.72}
  & \solverrank{3.16}{0.76} \\
\solverlabel{Acc.\ (\%) $\uparrow$}
  & 56.17 & 54.54 & 53.78 & 55.30 & 53.79 & 46.42 \\
\bottomrule
\end{tabular}

\endgroup
\end{table}

\section{Model Performance}
\begin{figure}[!htb]
    \centering
    \includegraphics[width=0.95\linewidth]{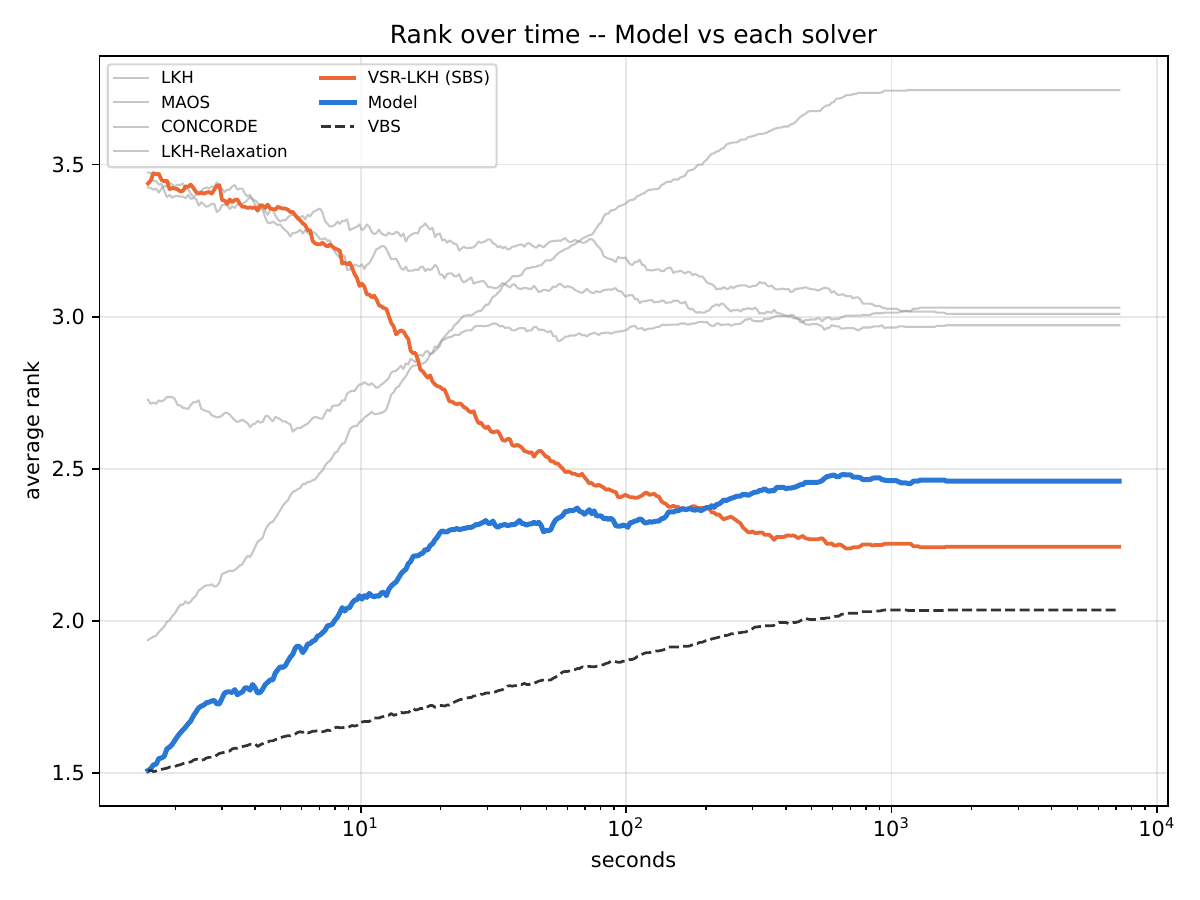}
    \caption{Average rank changes over time}
    \label{fig:rank}
\end{figure}

On average, ST-GNN-AAAS outperforms both the Single Best Solver (SBS) and the Random Forest (RF) baselines across all evaluated instances in terms of average rank and accuracy. 
However, when examining smaller intervals within the full prediction window, ST-GNN-AAAS does not always outperform the SBS as seen in Figure~\ref{fig:rank} and Table~\ref{tab:window_results}.
Notably, our ST-GNN-AAAS's per-interval accuracy declines monotonically from 93.1\% in the earliest prediction window (time steps 100-150) to 71.5\% in the latest window (400-500), while the SBS's accuracy rises monotonically over the same intervals, from 30.5\% to 87.5\%. 
We attribute this reversal to two potential factors. 
First, the SBS solver VSR-LKH exhibits the characteristic anytime profile: it is rarely competitive within the first few hundred seconds but becomes increasingly likely to be the SBS as the time budget grows. 
However, our ST-GNN-AAAS is only given an early history window before VSR-LKH shows any sign of this later advantage, therefore it has no direct evidence in its input to favor VSR-LKH in the late stages. Second, and more fundamentally, during training we use a pairwise ranking loss following \citet{wang2018lambdaloss}, generalizing the metric-weighted pairwise formulation of LambdaRank by \citet{burges2010ranknet} by substituting the relative cost gap between solvers for the conventional NDCG-based weighting. 
Therefore, the weight is relatively large in early stages before solvers have converged, and shrinks towards zero when they have similar objective values. 
Therefore, the gradients assigned to later stages become smaller, which may contribute to the observed performance. 
Future work could explore alternative loss functions that better account for performance at later stages.

\begin{table}[htbp]
\centering
\caption{Performance across different interval of timesteps on the full test set. Lower average rank is better.}
\label{tab:window_results}
\small
\setlength{\tabcolsep}{4pt}
\renewcommand{\arraystretch}{1.15}

\begin{tabular}{lrrrrrrr}
\toprule
\textbf{Window} &
$n$ &
\multicolumn{3}{c}{\textbf{Average Rank}} &
\multicolumn{3}{c}{\textbf{Accuracy} (\%)} \\
\cmidrule(lr){3-5}
\cmidrule(lr){6-8}
&
&
Model &
SBS &
VBS &
Model &
SBS &
VBS \\
\midrule
ALL
& 267
& $2.26 \pm 0.69$
& $2.61 \pm 0.58$
& $1.85 \pm 0.79$
& $76.59 \pm 29.22$
& $65.49 \pm 32.52$
& 100.0 \\

100--150
& 267
& $1.70 \pm 0.85$
& $3.40 \pm 0.75$
& $1.56 \pm 0.81$
& $93.09 \pm 20.11$
& $30.49 \pm 44.90$
& 100.0 \\

150--200
& 267
& $2.00 \pm 0.96$
& $3.18 \pm 0.88$
& $1.65 \pm 0.83$
& $83.96 \pm 30.81$
& $39.36 \pm 46.14$
& 100.0 \\

200--300
& 267
& $2.31 \pm 0.95$
& $2.60 \pm 0.82$
& $1.79 \pm 0.82$
& $73.78 \pm 39.90$
& $59.87 \pm 44.61$
& 100.0 \\

300--400
& 267
& $2.42 \pm 0.78$
& $2.31 \pm 0.79$
& $1.97 \pm 0.82$
& $72.57 \pm 42.04$
& $79.65 \pm 35.93$
& 100.0 \\

400--500
& 267
& $2.46 \pm 0.74$
& $2.24 \pm 0.83$
& $2.04 \pm 0.82$
& $71.49 \pm 45.09$
& $87.53 \pm 32.96$
& 100.0 \\
\bottomrule
\end{tabular}
\end{table}

\subsection{Additional Ablation Results}
\label{app:ablation}
To evaluate the contribution of the dynamic tour features described in Section~\ref{Sec2.2}, we remove $d_{\mathrm{prev}}$ and $d_{\mathrm{next}}$ while retaining only the normalized $(x,y)$ coordinates as node features. The full results are reported below together with those of the complete model. Table~\ref{tab:ablation_tour_a} compares average rank and selection accuracy, while Table~\ref{tab:ablation_tour_b} reports the Wilcoxon $p$-values and instance-level win/tie/loss results against the SBS.
\begin{table}[H]
\centering
\caption{Detailed ablation comparison of ST-GNN-AAAS with and without dynamic tour features in terms of average rank and accuracy. SBS and VBS are shared across the two variants and are therefore reported once for each dataset.}
\label{tab:ablation_tour_a}


\footnotesize
\renewcommand{\arraystretch}{1.35}
\setlength{\tabcolsep}{1.2pt}

\resizebox{\linewidth}{!}{%
\begin{tabular}{@{}llrcccccccc@{}}
\toprule
\textbf{Source} & \textbf{Variant} & \textbf{$n$} &
\multicolumn{4}{c}{\textbf{Average Rank}} &
\multicolumn{4}{c}{\textbf{Accuracy (\%)}} \\
\cmidrule(lr){4-7}
\cmidrule(lr){8-11}
& & &
\textbf{Model} &
\textbf{SBS} &
\textbf{VBS} &
\textbf{Closed (\%)} &
\textbf{Model} &
\textbf{SBS} &
\textbf{VBS} &
\textbf{Closed (\%)} \\
\midrule

NAT. & Full & 8 &
$2.51 \pm 0.49$ &
$3.01 \pm 0.52$ &
$1.72 \pm 0.75$ &
38.76 &
$60.62 \pm 30.49$ &
$46.19 \pm 41.21$ &
100 &
26.82 \\

& w/o Tour & &
$2.59 \pm 0.44$ &
&
&
32.56 &
$55.09 \pm 32.27$ &
&
&
16.54 \\

\midrule

TNM & Full & 126 &
$1.81 \pm 0.65$ &
$2.39 \pm 0.63$ &
$1.39 \pm 0.60$ &
58.00 &
$69.95 \pm 32.01$ &
$52.22 \pm 30.97$ &
100 &
37.11 \\

& w/o Tour & &
$2.25 \pm 0.95$ &
&
&
14.00 &
$58.95 \pm 31.79$ &
&
&
14.09 \\

\midrule

TSPLIB & Full & 70 &
$2.79 \pm 0.36$ &
$2.96 \pm 0.43$ &
$2.58 \pm 0.62$ &
44.74 &
$90.59 \pm 19.15$ &
$85.28 \pm 28.58$ &
100 &
36.07 \\

& w/o Tour & &
$2.79 \pm 0.44$ &
&
&
44.74 &
$90.46 \pm 21.38$ &
&
&
35.19 \\

\midrule

VLSI & Full & 63 &
$2.52 \pm 0.44$ &
$2.62 \pm 0.39$ &
$1.98 \pm 0.62$ &
15.63 &
$76.35 \pm 27.18$ &
$72.50 \pm 24.60$ &
100 &
14.00 \\

& w/o Tour & &
$2.77 \pm 0.74$ &
&
&
-23.44 &
$67.13 \pm 31.30$ &
&
&
-19.53 \\

\midrule

ALL & \textbf{Full} & 267 &
\textbf{$2.26 \pm 0.69$} &
$2.61 \pm 0.58$ &
$1.85 \pm 0.79$ &
\textbf{46.05} &
\textbf{$76.59 \pm 29.22$} &
$65.49 \pm 32.52$ &
100 &
\textbf{32.16} \\

& w/o Tour & &
$2.52 \pm 0.82$ &
&
&
11.84 &
$69.02 \pm 32.01$ &
&
&
10.23 \\

\bottomrule
\end{tabular}%
}
\end{table}
\begin{table}[H]
\centering
\caption{Statistical comparison of ST-GNN-AAAS with and without dynamic tour features against the SBS.}
\label{tab:ablation_tour_b}

\small
\renewcommand{\arraystretch}{0.5}
\setlength{\tabcolsep}{10pt}

\makebox[\textwidth][c]{%
\begin{tabular}{llr c ccc}
\toprule
\textbf{Source} &
\textbf{Variant} &
\textbf{$n$} &
\textbf{$p$-value} &
\multicolumn{3}{c}{\textbf{Model vs. SBS (\%)}} \\
\cmidrule(lr){5-7}

& & & &
\textbf{Win} &
\textbf{Tie} &
\textbf{Lose} \\
\midrule

NAT. & Full & 8 &
$4.69 \times 10^{-2}$ &
62.5 & 25.0 & 12.5 \\

& w/o Tour & 8 &
$7.81 \times 10^{-2}$ &
50.0 & 25.0 & 25.0 \\

\midrule

TNM & Full & 126 &
$7.14 \times 10^{-11}$ &
69.0 & 10.3 & 20.6 \\

& w/o Tour & 126 &
$3.21 \times 10^{-2}$ &
57.1 & 8.7 & 34.1 \\

\midrule

TSPLIB & Full & 70 &
$1.32 \times 10^{-2}$ &
25.7 & 64.3 & 10.0 \\

& w/o Tour & 70 &
$1.85 \times 10^{-2}$ &
24.3 & 64.3 & 11.4 \\

\midrule

VLSI & Full & 63 &
$2.71 \times 10^{-2}$ &
50.8 & 20.6 & 28.6 \\

& w/o Tour & 63 &
$8.37 \times 10^{-1}$ &
42.9 & 19.0 & 38.1 \\

\midrule

ALL & \textbf{Full} & 267 &
$6.13 \times 10^{-13}$ &
\textbf{53.2} & 27.3 & 19.5 \\

& w/o Tour & 267 &
$2.02 \times 10^{-2}$ &
44.9 & 26.2 & 28.8 \\

\bottomrule
\end{tabular}%
}
\end{table}

\subsection{Selected Solver Solution Quality}
Following Eq.~\ref{eq:normalized_quality}, we evaluate
the solution quality of the selected solvers using
the mean normalized solution quality $\bar{n}$.
To compare performance relative to the virtual best
solver (VBS) and single best solver (SBS),
we define the relative quality gap as

\begin{equation}
\label{eq:relative_quality_gap}
\hat{m}_s =
\frac{\bar{n}_s - \bar{n}_{\mathrm{VBS}}}
{\bar{n}_{\mathrm{SBS}} - \bar{n}_{\mathrm{VBS}}},
\end{equation}

where $\bar{n}_s$ denotes the mean normalized
solution quality achieved by method $s$.
A lower $\hat{m}_s$ indicates better performance,
with $\hat{m}_{\mathrm{VBS}}=0$ and
$\hat{m}_{\mathrm{SBS}}=1$.
Values below 1 indicate better solution quality
than the SBS, while values above 1 indicate
worse performance.

The SBS is determined separately for each
instance family and prediction window.

Over the full prediction horizon, our method
achieves $\hat{m}_{\mathrm{ST-GNN-AAAS}}=0.2796$,
corresponding to a quality gap closure of
$72.04\%$.

\begin{table}[H]
\centering
\caption{Relative solution quality ($\hat{m}$) across TSP
instance families over prediction steps $[100,500)$.
Lower values indicate better performance.
Bold indicates the best non-oracle result.}
\label{tab:quality_family}

\small
\renewcommand{\arraystretch}{1.15}

\begin{tabular*}{0.94\linewidth}
{@{\extracolsep{\fill}}lccccc@{}}
\toprule
Method & Overall & NATIONAL & TNM & TSPLIB & VLSI \\
\midrule

\# Instances
& 267 & 8 & 126 & 70 & 63 \\

\# Evaluations
& 106800 & 3200 & 50400 & 28000 & 25200 \\

\midrule

VBS
& 0.0000 & 0.0000 & 0.0000 & 0.0000 & 0.0000 \\

\textbf{Ours}
& \textbf{0.2796}
& \textbf{0.1623}
& \textbf{0.3737}
& \textbf{0.1820}
& \textbf{0.2684} \\

\midrule

CONCORDE
& 1.0000 & 1.0000 & 1.0000 & 1.3426 & 1.0000 \\

VSR-LKH
& 2.5795 & 2.3855 & 3.1363 & 1.6513 & 2.8132 \\

MAOS
& 2.9959 & 1.0331 & 5.5337 & 1.0000 & 1.3195 \\

LKH-Relaxation
& 3.1357 & 2.1548 & 4.7362 & 1.2044 & 2.6502 \\

LKH
& 3.3361 & 1.9767 & 5.2337 & 1.0940 & 2.6885 \\

\bottomrule
\end{tabular*}

\par\vspace{3pt}
\begin{minipage}{0.94\linewidth}
\footnotesize
\emph{Note.}
The SBS is CONCORDE for Overall, NATIONAL, TNM,
and VLSI, and MAOS for TSPLIB.
The SBS is determined separately for each family.
The VBS and SBS correspond to $\hat{m}=0$
and $\hat{m}=1$, respectively.
\end{minipage}

\end{table}

\end{document}